\documentclass[runningheads]{llncs}
\usepackage{graphicx}
\usepackage{amsmath}
\usepackage{booktabs}
\usepackage{algorithm}
\usepackage{algpseudocode}
\usepackage{dblfloatfix}
\usepackage{enumerate}
\usepackage{amsmath,amssymb}
\usepackage[table,xcdraw, dvipsnames]{xcolor}
\usepackage{subcaption}
\usepackage{bm}
\usepackage{cite}
\usepackage{hyperref}

\begin{document}
\title{Instance-based Counterfactual Explanations for Time Series Classification}
%
%
\author{Eoin Delaney \inst{1,2,3} \and
Derek Greene \inst{1,2,3} \and
Mark T. Keane\inst{1,2,3}}
\authorrunning{E. Delaney et al.,}
%
\institute{School of Computer Science, University College Dublin, Dublin, Ireland \and
Insight Centre for Data Analytics, University College Dublin, Dublin, Ireland \and VistaMilk SFI Research Centre, Ireland\\
\email{\{eoin.delaney,derek.greene,mark.keane\}@insight-centre.org}}
\maketitle              
\begin{abstract} In recent years, there has been a rapidly expanding focus on explaining the predictions made by black-box AI systems that handle image and tabular data. However, considerably less attention has been paid to explaining the predictions of opaque AI systems handling \textit{time series} data. In this paper, we advance a novel model-agnostic, case-based technique -- \textit{Native Guide} -- that generates counterfactual explanations for time series classifiers. Given a query time series, \textit{$T_{q}$}, for which a black-box classification system predicts class, \textit{$c$}, a counterfactual time series explanation shows how \textit{$T_{q}$} could change, such that the system predicts an alternative class, \textit{$c'$}. The proposed instance-based technique adapts existing counterfactual instances in the case-base by highlighting and modifying discriminative areas of the time series that underlie the classification. Quantitative and qualitative results from two comparative experiments indicate that Native Guide generates plausible, proximal, sparse and diverse explanations that are better than those produced by key benchmark counterfactual methods.

\end{abstract}
\section{Introduction}
In recent years, the predictive success of machine learning systems has been undermined by their lack of interpretability and beset by growing public disquiet about the fairness, accountability, and transparency of intelligent systems \cite{gunning2019darpa, adadi2018peeking}. These challenges have led to major efforts in Explainable AI (XAI), where a raft of techniques has been developed to shed light on opaque predictions. Most of this research focuses on image and tabular data, with less attention being given to the explanation of time series data \cite{nguyenmodel}. Explaining time series predictions, arguably, presents a whole new set of issues for XAI, due to the multi-dimensional nature of the data, strong feature dependencies, and the need to define the contexts where explanations could be used. In this paper, we advance an explainable case-based reasoning (XCBR) solution to this XAI problem. 

Recently, a variety of CBR methods for XAI has been proposed (see \cite{schoenborn2020explainable} for a review).  For image and tabular data, these XCBR techniques provide factual, example-based explanations (e.g., \cite{kk2019survey, sormo_explanation_2005}), feature-importance explanations (CBR-LIME; \cite{recio2020cbr}), and counterfactual explanations \cite{keane2020good}. In particular, counterfactual explanations have become popular as a post-hoc explanation technique, with over 100 distinct methods being proposed \cite{keane2021if}. However, few of these methods consider the explanation of time series \cite{Karlsson2018, guidotti2020explainingtime, ates2020counterfactual, delaney2020instance}. Hence, we advance \textit{Native Guide}, a novel model-agnostic explanation technique for time series classification (TSC) systems that provides counterfactual explanations for their predictions. 

\begin{figure}[!t]
    \centering
    \includegraphics[width=0.7\textwidth]{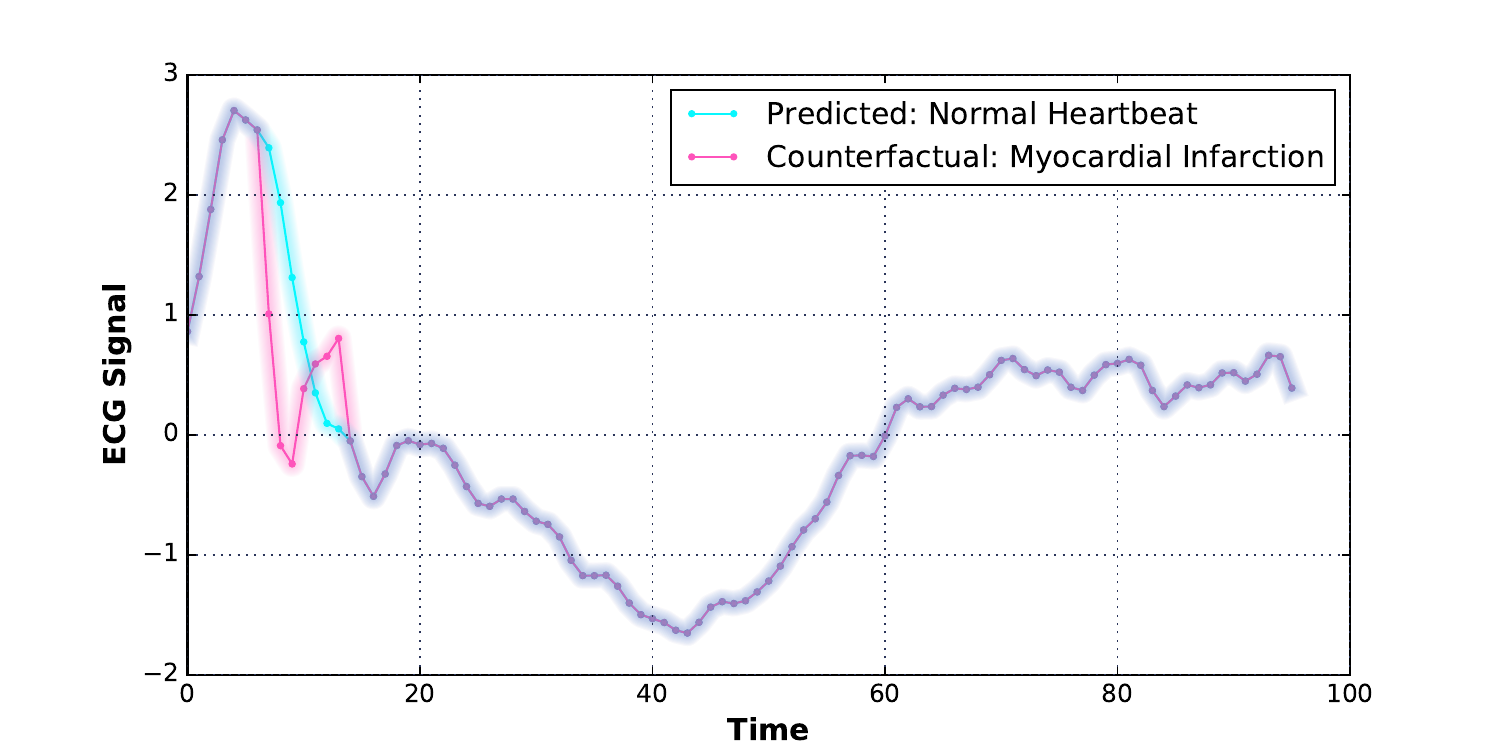}
    \caption{A counterfactual instance explains the classification of an ECG signal. Here, a black-box's classification of a normal heartbeat is explained with a counterfactual, from \textit{Native Guide}, showing an abnormal, heart-attack signal.}
    \label{fig:my_label}
\end{figure}

\textbf{XAI's promise for time series classification (TSC).} TSC has demonstrated significant promise in a variety of domains, including healthcare and food spectroscopy. However, there is a requirement to explain these decisions to end users. In healthcare, one practical application involves the classification of electrocardiogram signals, where explainable insights can aid medical practitioners in determining what portions of the time series are most informative for detecting abnormalities \cite{olszewski2001generalized} (e.g. myocardial infarction). Figure 1 shows one such example, where a cardiologist might be shown the normal heartbeat of a patient along with a counterfactual signal as an explanation, basically saying ``for this patient, their normal profile looks like this (purple-blue line), but if it changes to this counterfactual profile (purple-pink line), then they are experiencing an infarction" (see also Figure 2). Similar examples can be found in spectroscopy analyses when determining the provenance of different foods. For example, near-infrared spectrographs can distinguish between Arabica and Robusta coffee beans or honey from different regions \cite{briandet1996discrimination} (see also Figure 7). By identifying portions of the time series that are discriminative for classification, cheaper sensors can be designed that only consider a small portion of the wider spectra. Similarly, in deep learning systems, explainable insights can uncover the portions of a time series that may be most prone to adversarial attacks and show them to model developers \cite{fawaz2019adversarial}.
 
\begin{figure}[!t]
\begin{subfigure}{0.5\textwidth}
\includegraphics[width=0.9\linewidth, height=4cm]{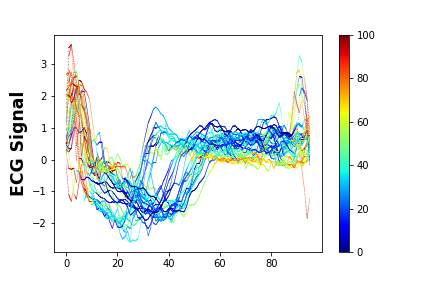} 
\caption{Normal Heartbeat}
\label{fig:subim1}
\end{subfigure}
\begin{subfigure}{0.5\textwidth}
\includegraphics[width=0.9\linewidth, height=4cm]{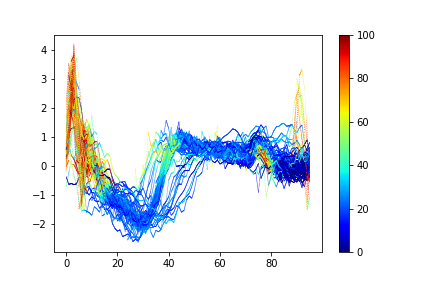} 
\caption{Myocardial Infarction}
\label{fig:subim2}
\end{subfigure}
\caption{Class Activation Maps (CAM) generated for the ECG200 dataset highlighting (in red) those areas of the time series which are most discriminative for a CNN Classifier. Here, the initial portion of the time series is most discriminative for both classes.}
\end{figure}

\textbf{Outline of paper.} In the remainder of this paper we first review the related work on time series XAI (Section 2). We then discuss the untapped promise of counterfactual explanations in TSC and the potential properties of good counterfactual explanations in this context (see Section 3). Next we describe the proposed technique (Section 4) and conduct comparative experiments to evaluate the quality of the explanations produced before discussing our results and suggesting promising avenues for future work (see Sections 5 \& 6).

\section{Related Work}

The XAI literature on explaining time series classification has progressed along similar lines to XAI, in general; initial techniques focused on explanation through visualization and feature-importance, rather than on instance-based methods, such as factual or counterfactual explanations \cite{lipton_2018, kenny_ICPR}.

\textit{Saliency} methods typically visualize an extracted explanation weight vector $\bm{\omega}$ that captures discriminative areas of a time series for classification \cite{nguyenmodel}. For example, Class Activation Maps (CAMs) \cite{zhou2016learning} utilize these weight vectors to highlight areas of a time series that are most informative for classification decisions of deep neural networks (DNNs) \cite{wang2017time, IsmailFawaz2019} (see Figure 2). Similarly, \textit{shapelets} can also find discriminative subsequences of a time series that can either be directly extracted from a set of time series \cite{ye2011time} or learned  by minimizing an objective function \cite{grabocka2014learning}. Shapelets can capture relationships between features and are closely related to saliency maps as both techniques offer visual explanations for classification tasks. Some have considered using shapelets for contrastive explanation \cite{guidotti2020explainingtime}. However, concerns have been raised about the interpretability of shapelets produced by the deployed \textit{learning-shapelets} algorithm \cite{wang2019learning}. 

\textit{Feature-importance analyses} are another method used to find relevant portions of a time series for use in explanation. Many state-of-the-art time series classifiers (e.g. Mr-SEQL \cite{LeNguyen2019a}) transform the input data and deploy a linear model for classification, where $\bm{\omega}$ can be directly extracted from the regression coefficients of the classifier. Indeed, model-agnostic techniques such as LIME \cite{ribeiro_why_2016} and SHAP \cite{lundberg2017unified}, can be used to compute $\bm{\omega}$ if it is not readily provided by the base classifier \cite{nguyenmodel}. However, concerns about the stability of these methods have been raised through examining how small perturbations can change the explanation \cite{adebayo2018sanity, nguyenmodel}. Schlegel et al. \cite{Schlegel2019} tested the informativeness and robustness of different feature-importance techniques in time series classification. LIME was found to produce poor results across all evaluated datasets (a problem attributed to the high dimensionality of the data); in contrast, saliency-based approaches and SHAP were found to be more robust across different architectures. 

More recently, a handful of \textit{instance-based techniques} have been proposed to explain time series classification. \textit{Prototypes} are instances that are maximally representative of a class and have demonstrated promise in producing global insights for time series classification in the healthcare domain \cite{Gee2019} but they do not provide insights into the most discriminative areas of the time series. Case-based approaches using twin systems \cite{kenny2019twin, kk2019survey, sani2017learning} have also been extended to time series data; Leonardi et al. \cite{leonardi2020deep} suggested mapping features from a DNN to a CBR system for interpretable haemodialysis classification. However, these techniques do not consider very popular counterfactual explanations. In an earlier unpublished version of the present paper \cite{delaney2020instance}, we considered how instances from the case-base could be retrieved for counterfactual explanation. However, we did not retrieve and integrate discriminative feature information in counterfactual generation, a significant novelty in the current method. Here, we advance a new XCBR method for generating good explanatory counterfactuals for any black-box time series classifier.

\section{Good Counterfactuals for Time Series: Key Properties}
There is a growing consensus that counterfactual explanations are causally informative \cite{lipton_2018, PearlMackenzie18}, psychologically effective \cite{miller2019explanation, Byrne2019, Dodge2019, keane2020good, molnar2020}, and legally compliant with respect to GDPR \cite{wachter2017counterfactual}. Arguably, counterfactuals provide more robust and informative explanations than feature-importance methods, such as LIME or SHAP \cite{Guidotti2019}. Although it can be difficult to visualize counterfactual explanations for tabular data  \cite{Mothilal2020}, in the time series domain their visualization is more straight-forward (see Figure 1). However, counterfactual XAI solutions for time series classification are rare (see e.g \cite{Karlsson2018, guidotti2020explainingtime, ates2020counterfactual} for closest works) and we know of no existing XCBR solutions. Indeed, it is unclear if (i) existing counterfactual techniques for tabular/image data can be applied to time series data, (ii) the properties of good counterfactuals from tabular and image data transfer to the time series domain. In Section 4, we present the details of our novel XCBR method, but before that we first consider four potential properties of good counterfactual explanations for time series: namely, proximity, sparsity, plausibility, and diversity.

\textbf{Proximity.}
Proximity refers to how close the to-be-explained query is to the generated counterfactual instance. Typically, closeness is measured using predefined distance metrics; close counterfactuals measured using Manhattan distance have been found to be informative \cite{Mothilal2020,keane2020good}. Following recent recommendations on evaluation \cite{keane2021if,downscruds,karimi2020modelcf}, we use several different distance metrics and a relative counterfactual distance measure, to monitor the proximity of the generated counterfactual with respect to existing in-sample counterfactual solutions \cite{keane2020good}.

\textbf{Sparsity.} As noted by \cite{Mothilal2020}, counterfactual instances that change fewer features are preferred for informative explanations. Keane and Smyth \cite{keane2020good} suggested that a sparsity of $\leq$2 feature differences was preferable for tabular data, on psychological grounds (that have been confirmed in recent user studies). However, the multi-dimensional nature of time series data means that a simple application of this idea is untenable. For image data, it has been argued that counterfactuals need to modify ``semantically-meaningful" features instead of small, pixel-level features that may not be humanly-perceptible  \cite{samangouei2018explaingan,kenny2020generating}. For time series data, it has been proposed that semantically-meaningful/discriminative information is contained in contiguous subsequences of the series \cite{ye2011time, LeNguyen2019a}. So, by analogy, we argue that ``good", sparse counterfactuals  need to modify a single discriminative portion of the time series (i.e., a contiguous subsequence), rather than distributed, discrete time-points in series (e.g., see Figures 1 \& 7).

\textbf{Plausibility.} Informative counterfactual explanations also need to be plausible \cite{miller2019explanation}. Many suggest that proximity is a good proxy for plausibility \cite{molnar2020}, though others argue that falling within the data distribution is a better proxy \cite{VanLooveren2019, kenny2020generating, Poyiadzi2020, laugel2019dangers}. Poyiadzi et al. \cite{Poyiadzi2020} argue that plausible counterfactuals are representative of the underlying data distribution. Figure 6 shows some examples of implausible counterfactuals that are out-of-distribution, even though they have high proximity. Hence, in our evaluations we explore several novelty-detection algorithms to find better measures of plausibility/implausibility (see e.g. \cite{kanamori2020dace}).

\textbf{Diversity.} Mothal et al. \cite{Mothilal2020} advanced the idea that a system should be able to produce multiple \textit{diverse} explanations for a single query case. One advantage of this is that different users may find different explanations helpful \cite{schoenborn2020explainable}. So, our proposed method generates multiple explanations for a single test instance. However, we explicitly ensure that diversity should not come at the cost of either (i) plausibility or (ii) the loss of semantically meaningful information. 

In the next section, we describe the proposed Native Guide method, before we consider a series of tests of it on several different datasets.

\section{Native Guide: Counterfactual XAI for Time Series}

\begin{figure}[h!]
    \centering
    \includegraphics[width=0.6\textwidth]{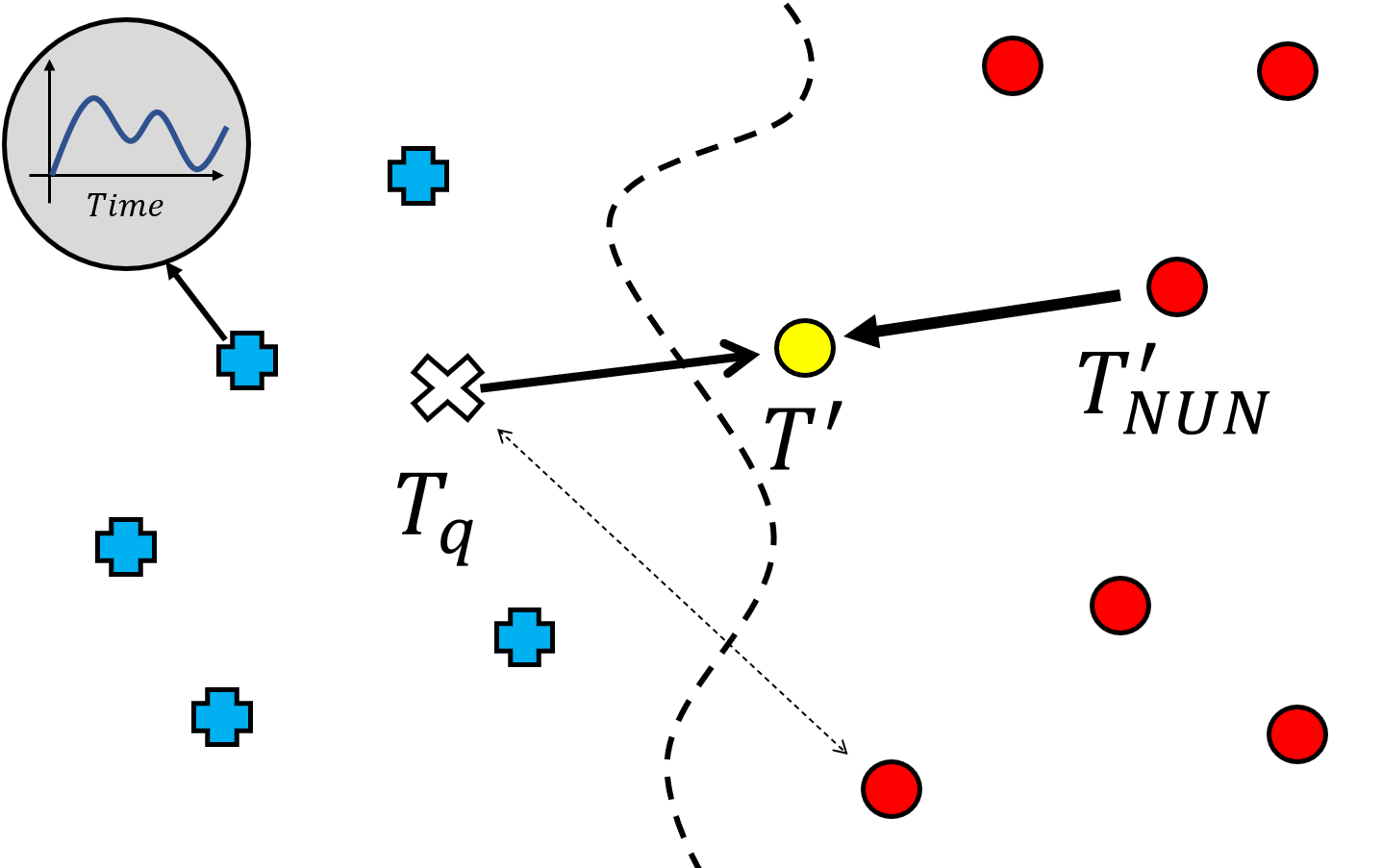}
    \caption{A query time series $T_{q}$ (X with solid arrow) and a nearest-unlike neighbor,  $T'_{NUN}$ (red circle with solid arrow) are used to guide the generation of counterfactual $T'$ (see yellow circle) in a binary classification task. Another in-sample counterfactual (i.e., the \textit{next} NUN; other red circle with dashed arrow) could also be used to generate another counterfactual for diverse explanations.}
    \label{fig:my_label}
\end{figure}

Like other case-based XAI methods \cite{kenny2019twin, keane2020good, Nugent2009, leake2005introduction}, at its core Native Guide relies upon existing instances in the training data, so-called \textit{native guides} or nearest unlike neighbors (NUNs), that it retrieves and adapts to generate counterfactual explanations (see Figure 3). In this section, we outline the two main steps in the algorithm, after first describing the notation adopted.

\textbf{Notation.} Staying consistent with the notation of \cite{goyal2019counterfactual, guidotti2020explainingtime}, 
a time series $T = \{<t_1,t_2,...,t_m>\}$ is an ordered set of real values, where $m$ is the length. A time series data set \textbf{T} $= \{T_1,T_2...,T_n\} \in \mathbb{R}^{n \times m}$ is a collection of such time series where each time series has a class label $c$ forming a vector of class labels $\textbf{Y} \in \mathbb{Z}$. Consider a black-box classifier $b(T)$ that takes a time series $T$ as an input and predicts a probability output $P(\textbf{Y}|T)$ over the label output space. Given a to-be-explained query time series $T_{q}$, with predicted label $c$ from the black-box classifier (formally $b(T_{q}) = c$), a counterfactual explanation aims to find how $T_{q}$ needs to change for the system to classify it alternatively, as $c'$. We refer to $T'$ as a counterfactual explanation for $T_{q}$ such that $b(T') = c'$. Although there are many candidate solutions for $T'$, the method prioritizes those that meet the four key properties of proximity, sparsity, plausibility and diversity. 

\textbf{Step 1: Retrieve native guide.} Given a query time series, $T_{q}$, find a counterfactual instance, $T'_{Native}$, that exists in the case-base. An example of one such instance is the query's nearest unlike neighbor $(T'_{NUN})$. In using these ``native counterfactual" cases the method guarantees the explanation's \textit{plausibility} as it is, by definition, within the distribution. However, such instances are not guaranteed to be sufficiently proximate to the query or, indeed, sparse, so an adaption step is necessary to generate the ``explanatory counterfactual", $T'$ (see Figure 3).

\textbf{Step 2: Adapt native guide to generate counterfactual.}  To produce a more proximate explanatory counterfactual, $T'$, the native guide, $T'_{Native}$ is perturbed towards the to-be-explained query-case, $T_{q}$ (see Figure 3). Typically, counterfactual methods use some $L_{p}$ distance metric to guide this perturbation (such as Manhattan distance, \cite{wachter2017counterfactual}) and in time series where dynamic time warping (DTW) distance is often more appropriate an analogous averaging technique known as weighted dynamic barycentre averaging can be used \cite{Forestier2017}. In cases where we are explaining a deep-learner's predictions, the feature-weight vectors of the classifier, $\bm{\omega}$, can be used to perturb ``semantically-meaningful" features of the time series, rather than the ``raw" time series data, to guarantee sparsity\footnote{Note, SHAP can also be used to generate such vectors, if we are directly explaining any given model, rather than twinning.}. Accordingly, using the feature-weights, the method seeks to modify contiguous, subsequences, rather than the whole time series, as follows:
{\begin{itemize}
\begin{samepage}
\centering
    \item[] $T_{q} = \{<t_{1}, t_{2}, t_{3}, t_{4}, t_{5} ..., t{n}>\}$ s.t. $b(T_{q}) = c$
    \item[] $T' =\{<t_{1}, \textcolor{red}{t'_{2}}, \textcolor{red}{t'_{3}}, \textcolor{red}{t'_{4}}, t_{5} ..., t{n}>\}$ s.t. $b(T') = c'$
    
\end{samepage}
\end{itemize}}

\noindent Specifically, the feature-weight vector, $\bm{\omega}$, can be extracted using techniques such as Class Activation Mapping in the case of DNNs (see e.g. Figure 2). Given $T'_{Native}$ and \bm{$\omega$}, the most influential contiguous subsequence (measured by the magnitude of weights in \bm{$\omega$}) is identified and the corresponding region in $T_{q}$ is replaced with these values. This process can be initialized using a small subsequence and the length of this subsequence can be iteratively incremented until $b(T') = c'$. In the very worst case scenario, the size of the subsequence will be equal to the length of $T_{q}$ and the native counterfactual $T'_{Native}$ is returned. This adaptation step improves the \textit{proximity} and \textit{plausibility} of the generated counterfactuals. Finally, \textit{diversity} can also be met, as other in-sample instances can be used as guides (e.g., the \textit{next} nearest unlike neighbor), to produce alternative counterfactual explanations for the original query (see Figure 3).

\begin{table}
\centering
\caption{Summary of TSC datasets used to evaluate counterfactual explanations}
\begin{tabular}{|l||l|l|l|l|l|}
\hline
\multicolumn{1}{|l|}{\textbf{Dataset}} &
  \multicolumn{1}{l|}{\textbf{Train Size}} &
  \multicolumn{1}{l|}{\textbf{Test Size}} &
  \multicolumn{1}{l|}{\textbf{Length}} &
  \multicolumn{1}{l|}{\textbf{Type}} &
  \multicolumn{1}{l|}{\textbf{No.Classes}} \\ \hline
CBF       & 30  & 900 & 128 & Simulated & 3 \\
Chinatown & 20  & 343 & 24  & Traffic   & 2 \\
Coffee    & 28  & 28  & 286 & Spectro   & 2 \\
ECG200    & 100 & 100 & 96  & ECG       & 2 \\
GunPoint  & 50  & 150 & 150 & Motion    & 2 \\ \hline
\end{tabular}
\end{table}

\section{Testing Native Guide: Two Comparative Experiments}
We test the Native Guide counterfactual method in two experiments evaluating how it meets the properties of good explanatory counterfactuals relative to two benchmark methods on 5 representative datasets. Our focus is on explaining a black-box fully convolutional neural network classifier (FCN). Experiment 1 assesses the proximity and sparseness of the counterfactuals generated. Experiment 2 examines the plausibility and diversity of the counterfactuals generated. Here, we describe the setup for these experiments in terms of the datasets, comparative benchmark methods and black-box classification system.

\vskip 0.8em
\noindent\textbf{(I) Datasets.} Five diverse datasets (binary and multiclass) from the UCR archive \cite{Dau2019} (see Table 1) were used for the classification task. To encourage reproducibility we use the default train-test splits provided by the archive and provide all experimental code, fully detailing hyper-parameters\footnote{\url{https://github.com/e-delaney/Instance-Based_CFE_TSC}}. 

\vskip 0.8em
\noindent\textbf{(II) Baseline models.} The performance of Native Guide was compared to two baseline models: the $w$-counterfactual and NUN-CF methods. The \textit{w-counterfactual method} (\textit{w}-CF) proposed by Wachter et al., \cite{wachter2017counterfactual} is a key benchmark method; it is the most cited counterfactual XAI method in the literature and many other methods are variants of it\footnote{We tried and failed in these tests, to use DiCE \cite{Mothilal2020}, a variant of \textit{w}-CF with added constraints for diversity. We found that DiCE did not generate diverse counterfactuals within reasonable time-limits, suggesting that it is not well suited to high-dimensional time series data (even for shallower ANNs).} \cite{keane2021if}. It proposes that that counterfactuals can be generated by minimizing a loss function;
\begin{align}
    L(x,x',y',\lambda) = \lambda(b(x') - c')^2 + d(x,x')
 \end{align}
\begin{align}
    arg\underset{x'}{min}\underset{\lambda}{max} L(x,x',c',\lambda)
\end{align}
The first collection of terms in this loss function encourage the output of the classifier \textit{b}, to be close to the desired class $c'$. The $\lambda$ parameter acts as a balancing term. The distance metric $d(x,x')$ measures the amount of change between the to-be explained instance $x$ and the counterfactual candidate $x'$. A Manhattan distance weighted feature-wise with the inverse median absolute deviation (MAD) is typically used here in order to ensure the generation of sparse solutions that are robust to outliers \cite{wachter2017counterfactual}. One noted weakness of the $\lambda$ parameter is that it tends to infinity raising stability issues in counterfactual generation \cite{russell2019efficient}. The  second method used, the \textit{NUN-CF method}, can be viewed as a simplified variant of Native Guide; it simply uses the NUN for the query case directly, without any adaptation of deep or discriminative features (e.g., see \cite{Nugent2009}). This model represents a good comparison point as it allows us to see the contributions of the adaptation steps in Native Guide.

\vskip 0.8em
\noindent\textbf{(III) Time series classifier.} The black-box classifier used was a fully convolutional neural network (FCN)\footnote{Counterfactuals for other classifiers, such as MR-SEQL, were found but not reported.}, by \cite{wang2017time}, a state of the art DNN architecture for time series classification (Figure 4). Notably, the Global Average Pooling (GAP) layer reduces the number of parameters in a neural network while enabling the use of the Class Activation Map (CAM) \cite{zhou2016learning}; the latter highlights parts of the input time series that contribute the most to a given classification, enabling the extraction of $\bm{\omega}$. For each test/query instance, counterfactuals were generated using (i) \textit{NUN-CF}, where the NUN, using a Euclidean distance measure, was selected \cite{nugent_case-based_2005}, (ii) \textit{w-Counterfactual} method (\textit{w}-CF), as proposed by Wachter et al. \cite{wachter2017counterfactual}, initializing it with $\lambda =0.1$ and termination condition $P(c'|T) \geq 0.5$, minimizing the loss function in (see EQ1) with adaptive Nelder-Mead optimization (iii) \textit{Native Guide}, using the closest in-sample counterfactual and the feature-importance vector, $\bm{\omega}$, given by the Class Activation Map (CAM) \cite{nguyenmodel}.

\begin{figure}
    \centering
    \includegraphics[width=0.67\textwidth]{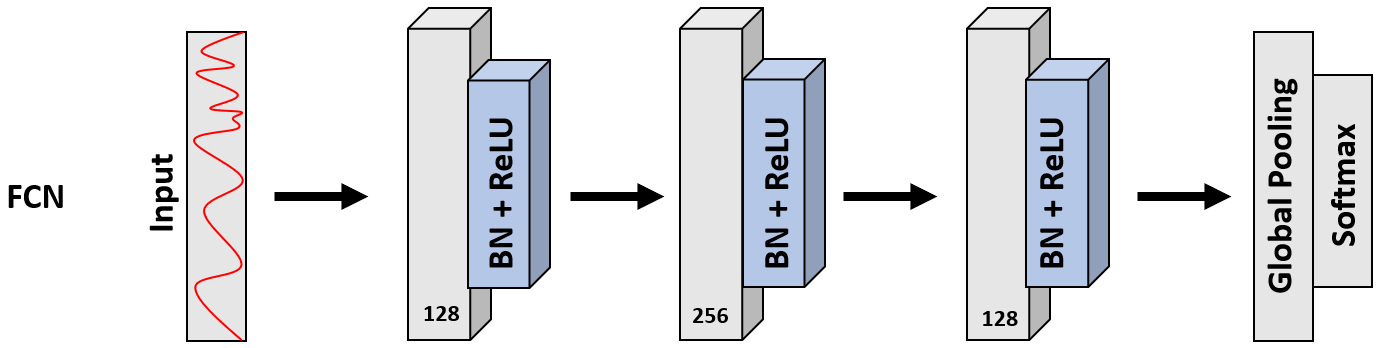}
    \caption{A fully convolutional neural network (FCN) with three convolutional layers, batch normalization, ReLu activations and global average pooling preceding the final softmax layer enabling the use of a Class Activation Map (CAM) \cite{wang2017time}.}
    \label{fig:my_label}
\end{figure}
\subsection{Experiment 1: Probing Proximity and Sparsity} 

This experiment compares the three counterfactual techniques on the five datasets, evaluating the counterfactuals produced in terms of proximity and sparsity. Proximity was evaluated using the relative counterfactual distance ($RCF$ =  $\frac{d(T_{q}, T')}{d(T_{q}, T_{NUN})}$) enabling explicit comparisons to in-sample counterfactual instances (as suggested by \cite{keane2020good,keane2021if}). Basically, this measure determines whether the distance between the query and the generated counterfactual is closer than that between the query and its ``naturally-occurring" NUN. As in some other studies \cite{karimi2020modelcf}, three distance metrics were used; (i)  Manhattan Distance ($\ell_{1}$ norm), (ii) Euclidean Distance ($\ell_{2}$ norm), and (iii) Chebyshev Distance ($\ell_{\infty}$ norm). 

\begin{figure}[h]
\begin{subfigure}{0.5\textwidth}
\includegraphics[width=1\linewidth, height=4.5cm]{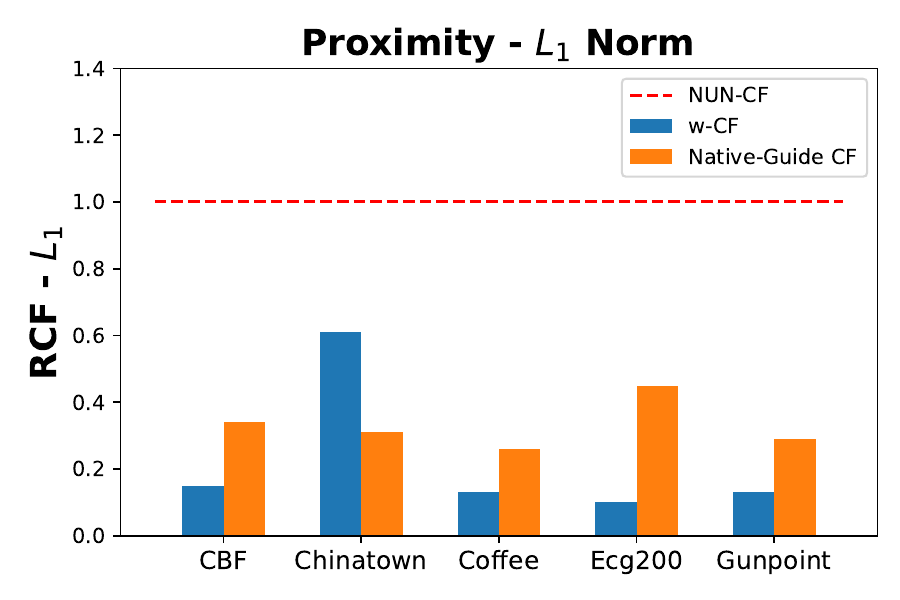} 
\caption{$L_{1}$}
\label{fig:subim1}
\end{subfigure}
\begin{subfigure}{0.5\textwidth}
\includegraphics[width=1\linewidth, height=4.5cm]{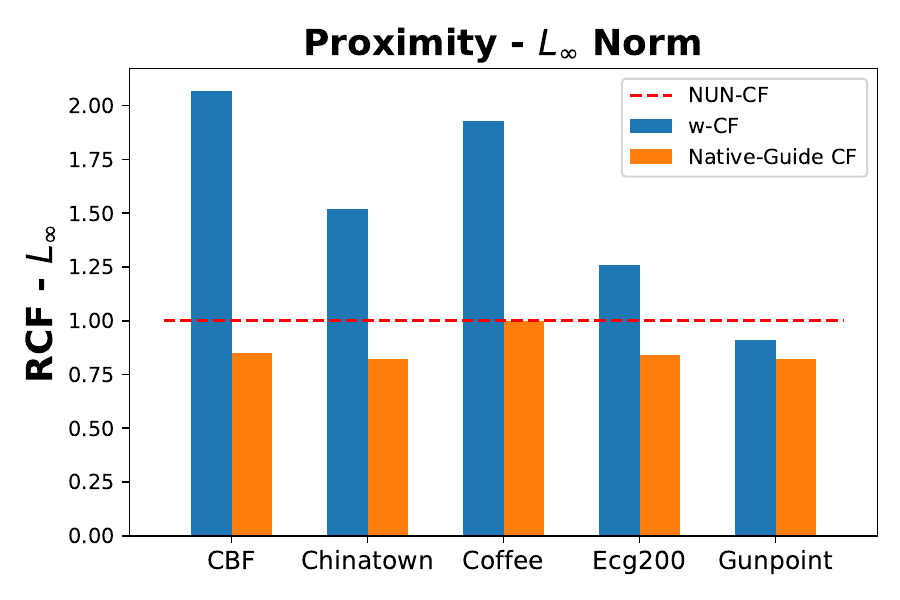}
\caption{$L_{\infty}$}
\label{fig:subim2}
\end{subfigure}

\caption{A comparison of the proximity of query-counterfactual pairs relative to query-NUN pairs for five datasets. In (a) the generated counterfactual explanations are closer to the the query compared to the in-sample NUNs, in terms of $\ell_{1}$ distance. Perhaps more interesting is the fact that the \textit{w}-counterfactuals are consistently less close than the NUNs, in terms of $\ell_{\infty}$ norm. This effect may be due to erroneous spikes in the counterfactual explanations generated by this method.}
\label{fig:image2} 
\end{figure}

\begin{figure}[h!]
\centering
\includegraphics[width=1\linewidth]{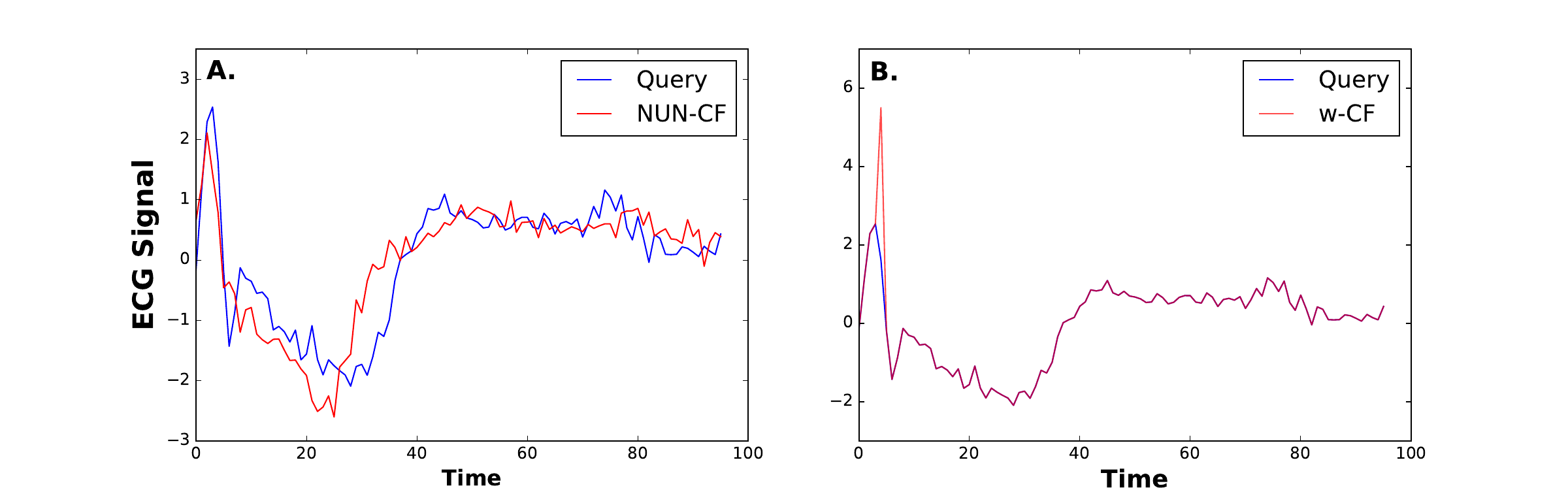} 
\caption{Comparing counterfactuals (red line) for an ECG200 classification (blue line) generated by  (a) NUN-CF and (b) \textit{w}-CF. Here, NUN-CF fails to generate a proximate/sparse solution and \textit{w}-CF's erratic spikes raise concerns about whether the counterfactual is out-of-distribution (see Figure 1 for comparison).}
\end{figure}

\begin{figure}[!h]
\centering
\includegraphics[width=8.6cm]{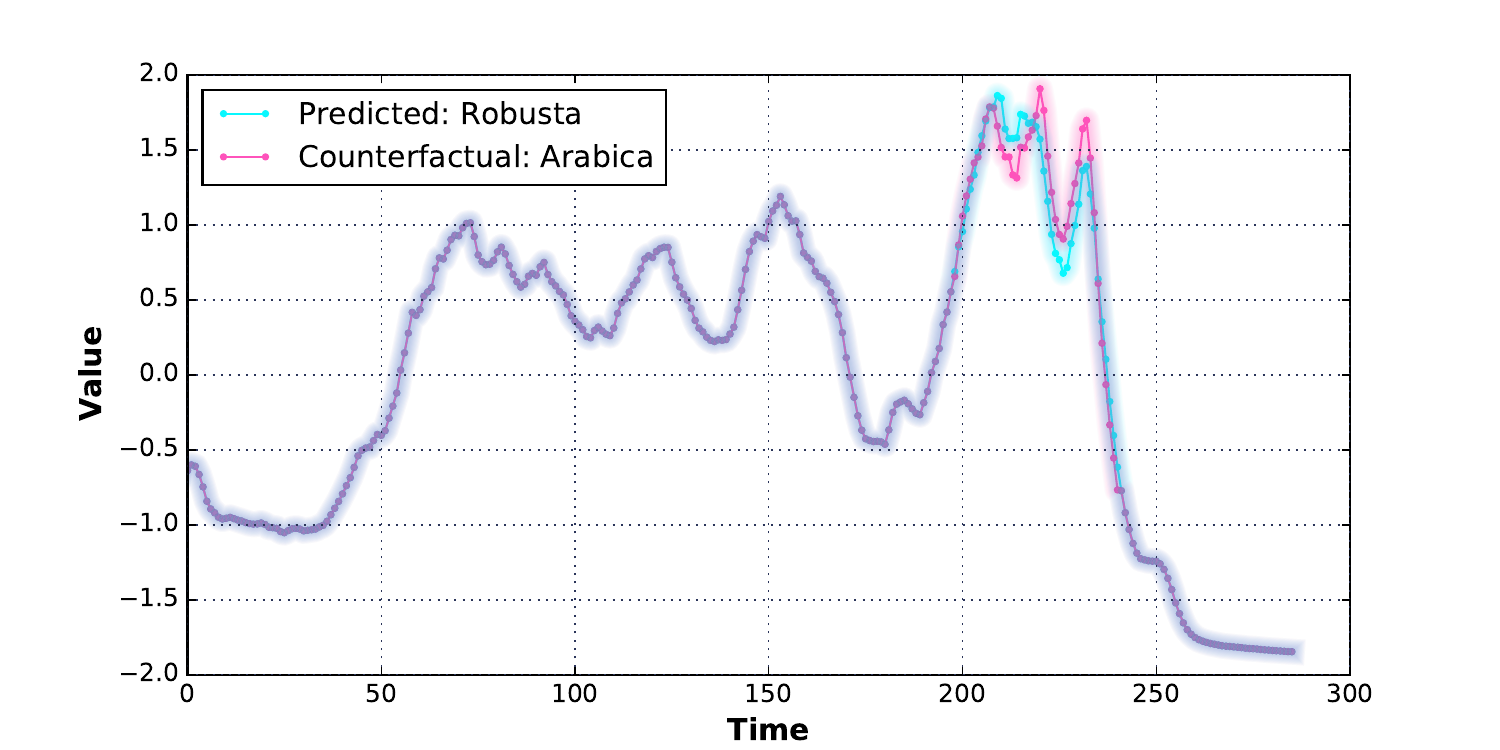} 
\caption{A Native Guide counterfactual explanation for the coffee dataset.  The method perturbs a contiguous subsequence corresponding to a semantically-meaningful and discriminative area of the spectrograph; this area provides information about the caffeine content of the coffee beans. Arabica coffee beans have a lower caffeine and chlorogenic acid content contributing to their finer taste and higher market value \cite{briandet1996discrimination}.}
\end{figure} 


\subsubsection{Results and discussion.}
Both Native Guide and the \textit{w}-CF counterfactual explanations produce proximate explanations that are significantly closer to the query instance compared to the existing NUNS, on both $\ell_{1}$ and $\ell_{2}$ norms (Wilcoxon, p$<$0.01). In the case of \textit{w}-CF this is somewhat unsurprising as it minimizes an $\ell_{1}$-based distance-metric optimizing to generate close, sparse counterfactuals. \textit{w}-CF is known to sometimes produce implausible counterfactual explanations \cite{keane2020good, kenny2020generating}. It is interesting to find that many of the perturbed features in its counterfactuals  can be erratic spikes in the time series, reflecting out-of-distribution occurrences (see Figure 6b). Moreover, the explanations produced by \textit{w}-CF often perturb several different features in non-contiguous locations of the time series, considering these values to be independent. Conversely, Native Guide constrains its perturbations to selected, contiguous subsequences producing counterfactual explanations that are more plausible and more meaningful (see e.g. Figures 1 and 7). These results indicate that good counterfactual explanations in the time series domain are not necessarily instances that are closest to the query, reflecting previous findings by Downs \textit{et al} for tabular data \cite{downscruds}. Notably, the $\ell_{\infty}$ norm seems to be able to diagnose counterfactual instances with erratic feature values. Counterfactual explanations produced by Native Guide are more proximate in terms of $\ell_{\infty}$ norm, further suggesting that the \textit{w}-counterfactuals may not be realistic. Admittedly, a more robust evaluation of plausibility should be considered when evaluating these methods. We turn to this issue in Expt. 2. 

\subsection{Experiment 2: Exploring Plausibility and Diversity}
In this experiment, we aim to evaluate the \textit{plausibility} of generated counterfactual explanations in time series using novelty detection algorithms to detect out-of-distribution (OOD) explanations. We implement the Local Outlier Factor Method \cite{breunig2000lof, kanamori2020dace}, Isolation Forest (IF) \cite{liu2008isolation} and OC-SVM \cite{scholkopf2001estimating} (on both raw time and matrix profile \cite{yeh2016matrix} representations of the time series). We test if Native Guide can generate \textit{diverse} explanations, when it uses alternative counterfactual instances as guides (see the other red instances shown in Figure 3). The datasets, classifier, and methods tested are identical to those in Experiment 1.

\begin{table}[!b]
\caption{Comparing the Native Guide (NG-CF) and \textit{w}-Counterfactual (\textit{w}-CF) models on plausibility using four OOD metrics (IF, LOF, OC-SVM, OC-SVM MP). Results indicate the percentage of generated counterfactuals that are out-of-distribution (n.b., lower scores are better and the best are highlighted in bold).}
\centering
\vskip 0.7em
\begin{tabular}{c|c|c|c|c|c|c|c|c|}
\cline{2-9}
\multicolumn{1}{l|}{}                                            & \multicolumn{8}{c|}{\textbf{Fully Convolutional Neural Network}}                                                                                                                                                                                                                                                                                                                             \\ \cline{2-9} 
\multicolumn{1}{l|}{}                                            & \multicolumn{2}{c|}{\cellcolor[HTML]{C0C0C0}{\color[HTML]{000000} \textbf{IF}}}                                    & \multicolumn{2}{c|}{\cellcolor[HTML]{FFFFFF}{\color[HTML]{000000} \textbf{LOF}}} & \multicolumn{2}{c|}{\cellcolor[HTML]{C0C0C0}{\color[HTML]{000000} \textbf{OC-SVM}}}                                & \multicolumn{2}{c|}{\cellcolor[HTML]{FFFFFF}\textbf{OC-SVM MP}} \\ \hline
\rowcolor[HTML]{FFFFFF} 
\multicolumn{1}{|c|}{\cellcolor[HTML]{FFFFFF}\textbf{Dataset}}   & \cellcolor[HTML]{C0C0C0}{\color[HTML]{000000} w-CF} & \cellcolor[HTML]{C0C0C0}{\color[HTML]{000000} NG-CF}         & {\color[HTML]{000000} w-CF}             & {\color[HTML]{000000} NG-CF}           & \cellcolor[HTML]{C0C0C0}{\color[HTML]{000000} w-CF} & \cellcolor[HTML]{C0C0C0}{\color[HTML]{000000} NG-CF}         & w-CF                       & NG-CF                              \\ \hline
\rowcolor[HTML]{FFFFFF} 
\multicolumn{1}{|c|}{\cellcolor[HTML]{FFFFFF}\textbf{CBF}}       & \cellcolor[HTML]{C0C0C0}0.15                        & \cellcolor[HTML]{C0C0C0}\textbf{0.09}                        & {\color[HTML]{000000} 0.09}             & {\color[HTML]{000000} \textbf{0.00}}   & \cellcolor[HTML]{C0C0C0}{\color[HTML]{000000} 0.69} & \cellcolor[HTML]{C0C0C0}{\color[HTML]{000000} \textbf{0.50}} & 0.61                       & \textbf{0.34}                      \\ \hline
\rowcolor[HTML]{FFFFFF} 
\multicolumn{1}{|c|}{\cellcolor[HTML]{FFFFFF}\textbf{Chinatown}} & \cellcolor[HTML]{C0C0C0}{\color[HTML]{000000} 0.48} & \cellcolor[HTML]{C0C0C0}{\color[HTML]{000000} \textbf{0.37}} & {\color[HTML]{000000} 0.11}             & {\color[HTML]{000000} \textbf{0.00}}   & \cellcolor[HTML]{C0C0C0}{\color[HTML]{000000} 0.44} & \cellcolor[HTML]{C0C0C0}{\color[HTML]{000000} \textbf{0.07}} & 0.87                       & \textbf{0.22}                      \\ \hline
\rowcolor[HTML]{FFFFFF} 
\multicolumn{1}{|c|}{\cellcolor[HTML]{FFFFFF}\textbf{Coffee}}    & \cellcolor[HTML]{C0C0C0}{\color[HTML]{000000} 0.41} & \cellcolor[HTML]{C0C0C0}{\color[HTML]{000000} \textbf{0.37}} & {\color[HTML]{000000} 0.04}             & {\color[HTML]{000000} 0.04}   & \cellcolor[HTML]{C0C0C0}{\color[HTML]{000000} 0.25} & \cellcolor[HTML]{C0C0C0}{\color[HTML]{000000} \textbf{0.14}} & 0.43                       & \textbf{0.21}                      \\ \hline
\rowcolor[HTML]{FFFFFF} 
\multicolumn{1}{|c|}{\cellcolor[HTML]{FFFFFF}\textbf{ECG200}}    & \cellcolor[HTML]{C0C0C0}{\color[HTML]{000000} 0.28} & \cellcolor[HTML]{C0C0C0}{\color[HTML]{000000} \textbf{0.26}} & {\color[HTML]{000000} 0.22}             & {\color[HTML]{000000} \textbf{0.02}}   & \cellcolor[HTML]{C0C0C0}{\color[HTML]{000000} 0.50} & \cellcolor[HTML]{C0C0C0}{\color[HTML]{000000} \textbf{0.16}} & 0.44                       & \textbf{0.13}                      \\ \hline
\rowcolor[HTML]{FFFFFF} 
\multicolumn{1}{|c|}{\cellcolor[HTML]{FFFFFF}\textbf{Gunpoint}}  & \cellcolor[HTML]{C0C0C0}{\color[HTML]{000000} 0.23} & \cellcolor[HTML]{C0C0C0}{\color[HTML]{000000} \textbf{0.20}}          & {\color[HTML]{000000} \textbf{0.19}}    & {\color[HTML]{000000} 0.23}   & \cellcolor[HTML]{C0C0C0}{\color[HTML]{000000} 0.18} & \cellcolor[HTML]{C0C0C0}{\color[HTML]{000000} \textbf{0.11}} & 0.57                       & \textbf{0.3}                       \\ \hline
\end{tabular}
\end{table}

\subsubsection{Results and discussion.} The counterfactuals produced by the Native Guide are consistently more plausible than those generated by the benchmark, w-counterfactual method (\textit{w}-CF; see Table 2). Results also confirm the hypothesis that proximity to the query is a poor heuristic for plausibility. One possible reason why case-based solutions produce more plausible counterfactual explanations is that they are grounded in the training data echoing previous findings by Laugel \textit{et al} \cite{laugel2019dangers}. Unlike Native Guide, \textit{w}-CF fails to perturb discriminative, meaningful subsequences \cite{ye2011time, LeNguyen2019a}. It is also interesting to note that different novelty detection algorithms produce very different results. For example the local outlier factor method was considerably less sensitive than the kernel-based techniques in detecting OOD explanations (see Table 2). Unlike many blind perturbation techniques, Native Guide has the ability to generate diverse counterfactual explanations (see Figure 8). This is particularly useful because (i) different users may prefer different explanations \cite{schoenborn2020explainable} and (ii) counterfactual explanations can also help humans to identify meaningful regions for classification (of which there may be many) \cite{goyal2019counterfactual}. For example, in the electrocardiogram domain we hypothesize that retaining counterfactual cases could help cardiologists to identify abnormalities that are useful for future problem scenarios. While one can evaluate diversity by monitoring feature wise distances between counterfactuals \cite{Mothilal2020}, the generated explanations may fail to satisfy domain constraints. Indeed, extensive user testing with experts will be an important avenue for future evaluation as novelty detection can be an imperfect proxy for plausibility.

\begin{figure}[!t]%
\begin{subfigure}[h]{0.48\textwidth}
\includegraphics[width=\textwidth]{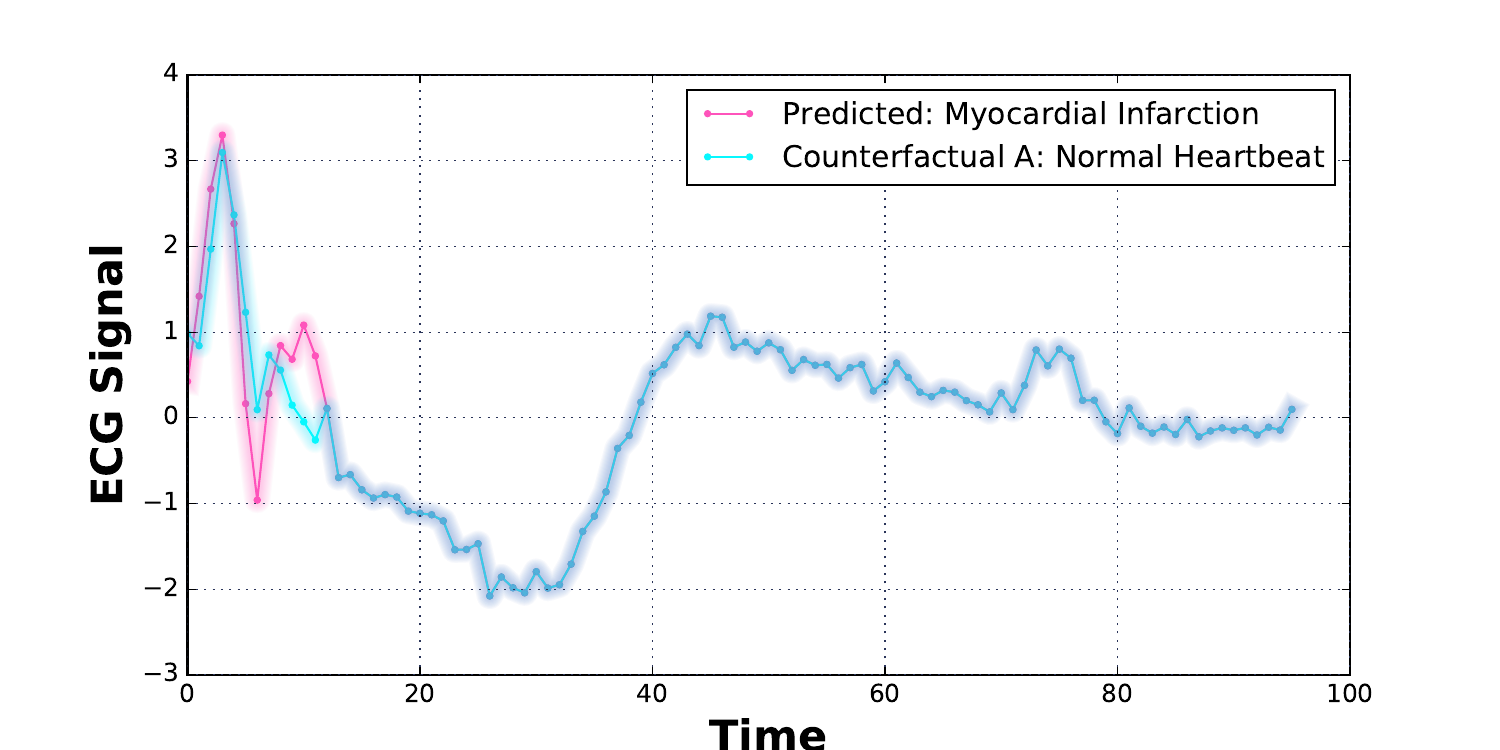}
\caption{Counterfactual A}
\end{subfigure}
\hfill\vrule\hfill
\begin{subfigure}[h]{0.48\textwidth}
\includegraphics[width=\textwidth]{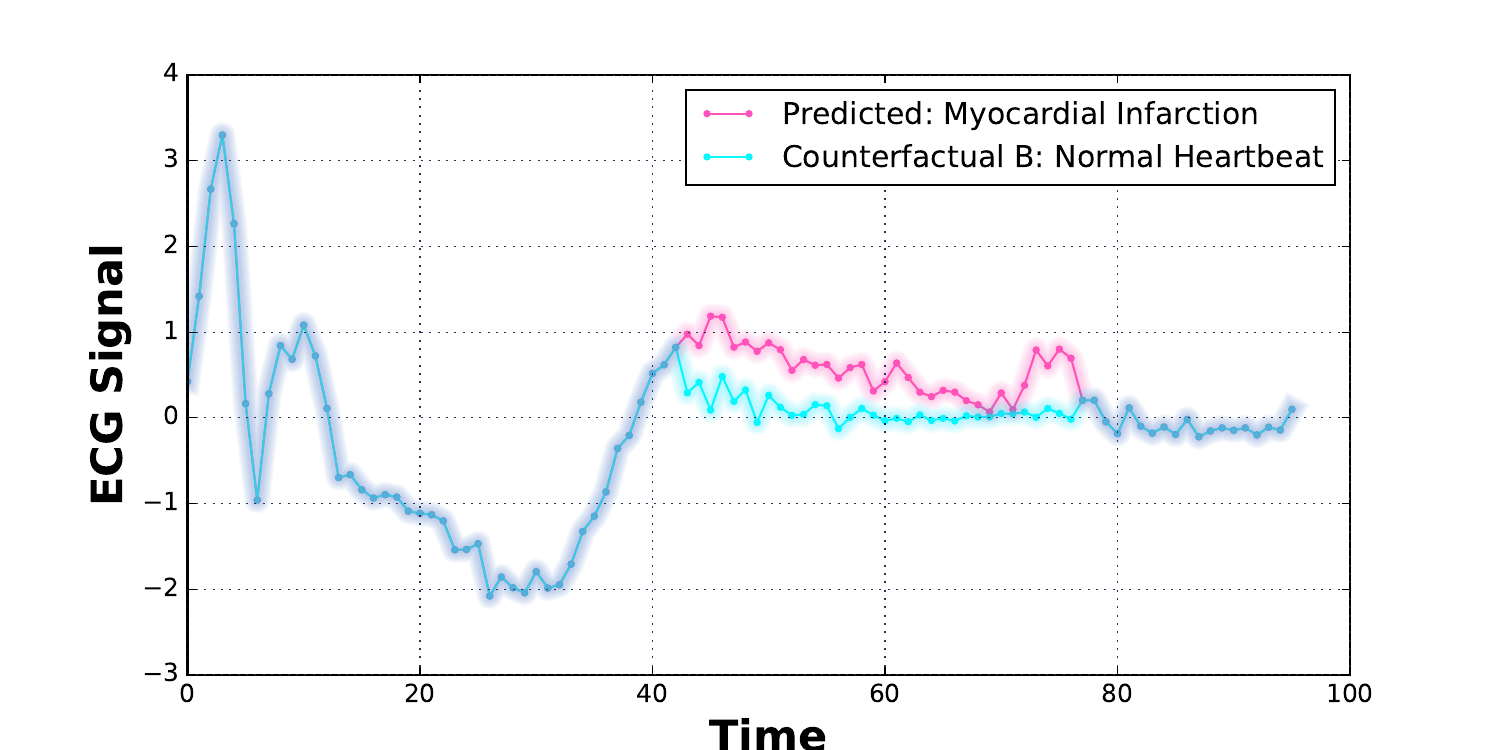}
\caption{Counterfactual B}
\end{subfigure}%
\caption{Two diverse counterfactual explanations, generated by Native Guide, for the same query case based on perturbing different in-sample counterfactual cases.}
\end{figure}

\section{Conclusion and Future Directions}
In this paper a novel case-based technique, \textit{Native Guide}, was proposed to provide proximate, sparse, plausible, and diverse counterfactual explanations for time series classification tasks. The method uses existing instances in the case-base to generate better counterfactual candidates. The technique is grounded in relevant evidence from the psychological and social sciences \cite{Byrne2019,miller2019explanation} and can integrate explanation weight-vectors extracted from techniques such as Class Activation Mapping \cite{zhou2016learning}. Comparative tests on diverse datasets from the UCR archive using a fully convolutional neural network, demonstrate that the explanatory counterfactuals produced by Native Guide are significantly better than (i) explanations that already existed in the case-base (from NUN-CF) and (ii) explanations produced by constraint-based optimisation techniques (from \textit{w}-CF). The experiments also indicated that techniques designed for tabular data often failed to produce meaningful explanations in the time series domain. Native Guide generates new time series data which holds promise for data augmentation purposes \cite{Forestier2017}. Given the ubiquitous nature of time series data and the frequent requirement for explanation, it is clear that experiments with human users and CBR solutions have much to offer in future work.

\subsubsection{Acknowledgements.} This publication has emanated from research conducted with
the financial support of (i) Science Foundation Ireland (SFI)
to the Insight Centre for Data Analytics under Grant
Number 12/RC/2289\_P2 and (ii) SFI and the Department of
Agriculture, Food and Marine on behalf of the Government
of Ireland under Grant
Number 16/RC/3835 (VistaMilk).
\bibliographystyle{splncs04.bst}
\bibliography{mybib.bib}

\end{document}